%% file: ms.tex
\documentclass{article}
\pdfoutput=1

\usepackage{microtype}
\usepackage{graphicx}
\usepackage{subfigure}
\usepackage{booktabs} 
\input{mathcommands.tex}

\usepackage{hyperref}

\usepackage[accepted]{icml2019}

\icmltitlerunning{Deep Q-Learning for Directed Acyclic Graph Generation}

\begin{document}

\twocolumn[
\icmltitle{Deep Q-Learning for Directed Acyclic Graph Generation}



\icmlsetsymbol{equal}{*}

\begin{icmlauthorlist}
\icmlauthor{Laura D'Arcy}{cardiff}
\icmlauthor{Padraig Corcoran}{cardiff}
\icmlauthor{Alun Preece}{cardiff}

\end{icmlauthorlist}

\icmlaffiliation{cardiff}{Cardiff University, Cardiff, Wales, CF10 3AT, UK}

\icmlcorrespondingauthor{Laura D'Arcy}{DArcyL@cardiff.ac.uk}

\icmlkeywords{Machine Learning, ICML, Directed Acyclic Graph, Reinforcement Learning}

\vskip 0.3in
]



\printAffiliationsAndNotice{}  

\begin{abstract}
We present a method to generate directed acyclic graphs (DAGs) using deep reinforcement learning, specifically deep Q-learning. Generating graphs with specified structures is an important and challenging task in various application fields, however most current graph generation methods produce graphs with undirected edges. We demonstrate that this method is capable of generating DAGs with topology and node types satisfying specified criteria in highly sparse reward environments. 
\end{abstract}

\section{Introduction}
\label{introduction_section}
Graph generation is a quickly growing area of study with applications in a wide range of problem domains, such as drug discovery and task scheduling problems \citep{RLDrugDiscovery,SchedulingQLearning}. 
Most existing graph generation methods use some form of supervised or semi-supervised learning requiring large amounts of training data. For example, \cite{MolGAN} use a data set of 133,885 molecules as a prior distribution for graph generation. However,  in certain application fields, such as distributed systems composition, prior examples are either sparse or nonexistent. We aim to create a method that can generate graphs with no prior data.

A few recent works have used reinforcement learning to generate undirected graphs where these methods require less training data or supervision. These works employ generative adversarial networks alongside proximate policy optimization \citep{RLDrugDiscovery}. Q-learning has also been implemented for graph construction, however it has been implemented with tabular methods and is therefore unusable at scale due to the exponential size of the state space that needs to be explored\citep{SchedulingQLearning}. 

We propose a novel deep Q-learning approach to construct directed acyclic graphs (DAGs). Deep Q-learning is a model-free reinforcement learning algorithm that is known to perform well with large action and state spaces that require function approximation. In this implementation, we combine Q-learning with a feed-forward graph convolutional neural network where actions correspond to the addition of a set of nodes and edges. This method can account for large scale directed DAGs, with multiple node types, and potentially continuously valued node features.

The rest of this paper is organized as follows. In Section~\ref{problem_section}, we provide an overview of the DAG generation environment and the characteristics of the DAGs relevant to the rest of this paper. Section~\ref{model_section} describes the architecture of the graph neural network that is used to map state action pairs to Q-values. Section~\ref{learning_section} describes how the Q-learning algorithm is applied. Section~\ref{results_section} provides preliminary results, before the conclusion in Section~\ref{conclusion_section}.

\section{Problem Definition}
\label{problem_section}

In this section we formulate the problem of DAG construction as a reinforcement learning problem. The goal of our learning agent is to construct a directed graph $\gG = (V,E)$, where each node is one of $b$ node types. The reinforcement learning problem is posed as an agent environment structure, where an agent interacts with an environment and receives numerical rewards. Representation of states, actions and rewards are defined below.

\begin{figure}[h!]
\begin{center}

\includegraphics[draft=false,width=0.9\linewidth]{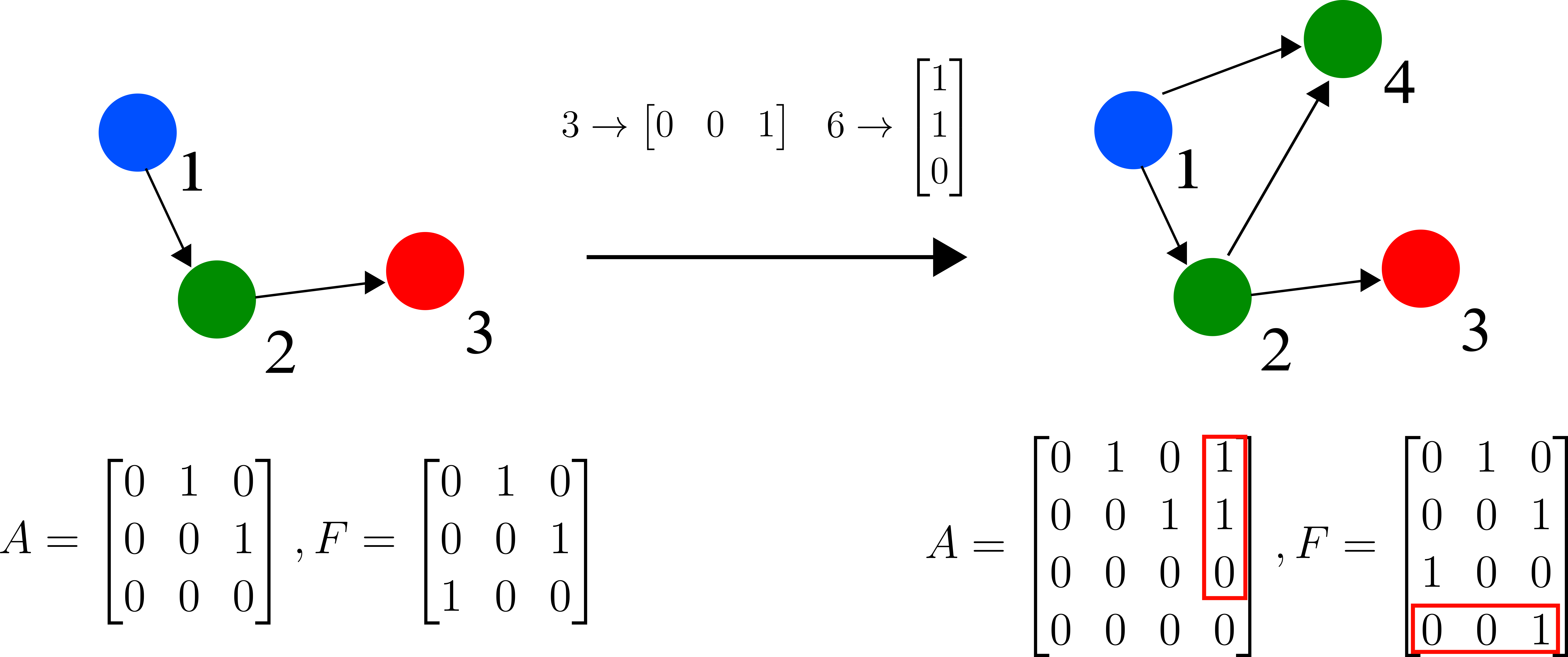}

\caption{An example of a single action in the DAG generation environment, along with the state descriptions before and after the action is taken. Here, the $4\textsuperscript{th}$ node is added, of type 3, with incoming edges from nodes 1 and 2. This action is described further in Section~\ref{action_subsection}.\vspace{-.4cm}}\label{action_figure}
\end{center}
\end{figure}

\subsection{State}
The state of the environment at time $t$ in a given episode corresponds to a DAG and is denoted $\gG_t$. The topology of $\gG_t$ is represented using a binary adjacency matrix $\mA$ where $\mA\left[i,j\right] = 1$ indicates that a directed edge exists from the node $i$ to the node $j$. Note that a 1 at the location $\left[i,j\right]$ in $\mA^\top$ indicates that a directed edge exists from the node $j$ to the node $i$.

Individual node types are represented using a feature matrix $\rmF \in \{0,1\}^{n\times b}$ containing a one hot encoding of the node types for each node, where $b$ is the number of possible node types and $n$ is the number of nodes in $\gG_t$. The initial state for each episode is the null graph.

\subsection{Action}\label{action_subsection}
In this environment, each action consists of two parts: adding a new node, and adding the set of incoming edges to that node. The new node $n_t$ can be any of the $b$ node types within the environment. This node type is added to the matrix of node features $\mF$ by appending the one hot encoding of this feature as an additional row. In Figure~\ref{action_figure}, a node of type 3 (represented as a green node) is added to the DAG, so a row of $\left[0,0,1\right]$ is added to the feature matrix of the DAG $\gG_t$. 

Next, the set of incoming edges $N_k^{in}$ for the new node is added. By only allowing for the addition of incoming edges, we create a topologically sorted graph, ensuring that all graphs created are acyclic. 

In order to select the set of incoming edges, all feasible extensions are encoded as a binary vector. In Figure~\ref{action_figure}, the possible vectors would range between 001, connecting only the $3\textsuperscript{rd}$ and $4\textsuperscript{th}$ nodes, and 111, connecting all previous nodes to the $4\textsuperscript{th}$ node. These vectors can then be converted from their binary format to an integer $i\in\left[1,2^{n-1}\right]$. This allows for enumeration of all feasible extensions when required to determine the action with the maximum Q-value. This also allows for selection of a random extension where performing exploration (as necessary in $\epsilon$-greedy Q-learning).

In Figure~\ref{action_figure}, the $4\textsuperscript{th}$ node is added, so the possible set of edges is in the range $i\in\left[1,7\right]$. In the figure, $i=6$, which converted to binary format is 110, so edges are added from the first and second nodes. This binary number is appended to the adjacency matrix of $\gG$ as a new column. Note that the null set of edges is excluded as a possible choice here: this is in order to prevent floating nodes, although this doesn't affect the theory of the method significantly. 

This method of adding a new node and multiple edges in a single action is in contrast to many graph generation methods which add a single node or edge per action \citep{RLDrugDiscovery}. This new method has the disadvantage of creating a much larger action space of $b\times(2^{n-1}-1)$ possible actions, as opposed to $b+{n-1}$ possible actions with a single node or edge addition. However, this method reaches a final DAG in fewer timesteps and accounts for symmetries in edge additions: when adding edges one at a time, a policy could develop which favours a particular order of adding edges to the set. As order of edge additions does not affect the final structure of the DAG, this asymmetrical policy is prevented, which is why adding an entire set of edges is preferable overall (see figure \ref{fig:actions}).

\subsection{Reward}

Reward design for implementation of this method depends largely on the application's problem domain. For scheduling problems, for instance, the reward could be a numerical value based on the speed and quality of service of the resulting task schedule. These rewards can consist of both intermediate rewards and a singular reward once a generation episode terminates. 

For the purposes of evaluating the proposed graph generation method, in this work we performed simulations in which a positive reward was returned if the agent produced a DAG isomorphic to a ground truth DAG, and a reward of 0 otherwise. This is clearly an extreme case, as even for a 10 node graph with only 1 node type, there would be $1.018\mathrm{e}{+13}$ possible final states, of which approximately 1-5 produce a non-zero reward. However, in order to prove the generality of this method the isomorphic reward case was used for the results in this paper, and tested on smaller graphs. Potential work for extension to larger graphs is discussed in Section \ref{conclusion_section}.

\section{Model}
\label{model_section}
The Q function is a mapping from a state action pair to a real value indicating the predicted future reward for the pair in question. We learn this mapping using a policy network, which takes the state of the environment and outputs a single scalar value which can be treated as the Q-value. Our policy network is in effect a feed-forward graph convolutional network, similar to graphSAGE \citep{GraphSAGE}. A major difference here is this network accounts for direction of the edges, in a manner similar to that of the struc2vec++ method \citep{struc2vec++}.

The network will take the adjacency matrix and one hot encoded features of a given state $\gG_t$ as the input. The feed-forward architecture consists of two convolutional layers, followed by a non-linearity, then a pooling layer. The network propagates this one hot representation through two graph convolutional layers, each consisting of two steps. The first step is a concatenation of the feature representation of that node, the sum of features of outgoing neighbours of that node, and the sum of features of incoming neighbours of that node. Incoming and outgoing neighbour feature sums are found by performing matrix multiplication with the one hot feature representation and the adjacency matrix containing in-degrees, and the adjacency matrix containing out-degrees (which is simply the transpose here), respectively. The second step of the convolutional layer is to pass the concatenation through a simple non-linearity.

After the two convolutional layers, the representation is passed through a linear layer and is then passed to a pooling layer, which aggregates the individual node representations into a graph representation using a sum function. A sum aggregator is chosen here as opposed to a mean or max aggregator as it preserves both the ratio of node types, unlike the max aggregator, as well as differentiating between different scales of graphs, unlike the mean aggregator \citep{howpowerfularegraphnn}. This representation is then passed through a final linear layer and ReLU unit, and then a fully connected linear layer, producing a single scalar which is used as the Q-value for the input state, as discussed further in section \ref{learning_section}.

Formally, this network is described as follows. Given a DAG $\gG=(V,E)$ with adjacency matrix $\mA$ containing in degrees and $\mA^\top$ containing out degrees. Each node $v$ is initially represented by a vector $\vh_v^0$ which is a one hot encoding of the node type. Node $v$ has a set of neighbouring nodes $N_k^{in}$ connected with incoming edges, and a set of neighbouring nodes $N_k^{out}$ connected with outgoing edges. In every convolutional layer $l$ the representations of the current node, and sums of incoming and outgoing neighbour node representations $\vh_v^l$, $\vh_w^l$, $\vh_u^l$ are concatenated:
\begin{equation}
    \rmH^{l+1}=\text{CONCAT}(\rmH^l,\mA\rmH^l,\mA^\top\rmH^l)
\end{equation}

This concatenated representation is then passed through a non-linearity (eq.~\ref{non-liner}) where $\rmW$ and $\rmB$ are trainable matrices.  
\begin{equation}\label{non-liner}
    \rmH^{l+1}=\text{ReLU}(\mH^l\rmW^l+\rmB^l)
\end{equation}

After two convolutions with this method, the node representations are pooled as in eq.~\ref{pooling}, resulting in a scalar value. 
\begin{equation}\label{pooling}
  \rmH^{l+1}= \text{SUM}(\mH^l\rmW^l+\rmB^l)  
\end{equation}

This value is passed through a final non-linearity and then a fully connected linear layer to produce the final result.

\section{Learning and Inference}
\label{learning_section}

This method generates training data for the policy network using an $\epsilon$-greedy implementation of the DQN method. With probability $\epsilon$ a random action is selected from the range of possible actions, otherwise the approximate $Q_{\text{max}}$ value is selected.
After performing this action, the current $Q$-value for the state is evaluated using the policy network. Additionally, the $Q$-value is calculated by adding the reward to the Q-value of following the optimal policy thereafter, as shown in eq.~\ref{Qcalc}.
\begin{equation}\label{Qcalc}
 Q^\pi(S_t,A_t) = r + \gamma\max_{a}Q(S_{t+1},a)   
\end{equation}
Note that $\max_{a}Q(S_{t+1},a)$ in eq.~\ref{Qcalc} is calculated using a separate target network, which is not trained every step, but instead has network weights from the policy network copied over after a fixed number of episodes.
Taking the difference of these Q-values produces our training error
$\delta = Q(S_t,A_t)-Q^\pi(S_t,A_t)$, and the loss function is then simply the $\normltwo$ norm $\mathcal{L}=|| \delta || $. This is back-propagated through the network, and the Q-values produced by the network for each state are then closer to the true Q-values. This update occurs online, or after every time step, as opposed to in batches. While producing noisier results, this tends to result in a higher convergence rate \citep{batchorno}. After training is complete, a greedy method is used to generate the DAGs, whereby the action with the largest Q-value is selected at each step.

A second learning agent utilizing prioritized experience replay \citep{schaul2015prioritized} was also implemented; this agent updated on the current timestep and additionally on 30 timesteps that have previously been selected that are held in a  memory buffer. The actions replayed are selected at random, with a higher probability for selection for actions with high rewards, and actions more recently added to the memory buffer.

\section{Results}
\label{results_section}
\begin{figure}[tb!]
    \centering
    \includegraphics[width=.95\linewidth]{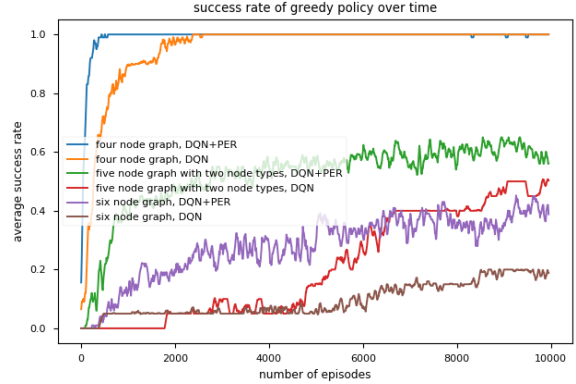}
    \caption{The average success rate of the greedy policy over time, for DAGs of  different sizes and different numbers of node types. For each graph type, a learning agent trains using an $\epsilon$-greedy policy over 10,000 episodes. The agent then runs an episode on the greedy policy after each $\epsilon$-greedy episode. Each learning agent is reset and trained 20 times, and the success rate of the greedy policy is the average of these 20 runs. For readability, the moving average over 50 episodes is plotted. Two types of agents are displayed: the standard DQN, and DQN with prioritized experience replay (DQN+PER).}
    \label{fig:success}
\end{figure}

\begin{table*}[h]
\caption{A comparison of average total reward for DAGs of varying size and number of unique node types, with basic DQN, DQN with prioritised experience replay, and an agent that selects actions at random. The rewards are totaled over 10,000 attempts, and averaged over 20 runs.}
\label{sample-table}
\vskip 0.15in
\begin{center}
\begin{small}
\begin{sc}
\begin{tabular}{lcccccc}
\toprule
  &\multicolumn{5}{c}{DAG Isomorphisms}\\
\midrule
DAG Sizes & 4 nodes & \multicolumn{3}{c}{5 nodes} & 6 nodes  \\
\midrule
Node Types & 1 & 1 & 2 & 3 & 1  \\
\midrule
Random    & 882&75&6& 2&0 \\
DQN    & 9524& 5242&2083& 1374&955  \\
{DQN+PER}   & 9902& 7380&5169& 796&2802         \\

\bottomrule
\end{tabular}
\end{sc}
\end{small}
\end{center}
\vskip -0.1in
\end{table*}

\begin{figure}[]
    \centering
    \includegraphics[width=.89\linewidth]{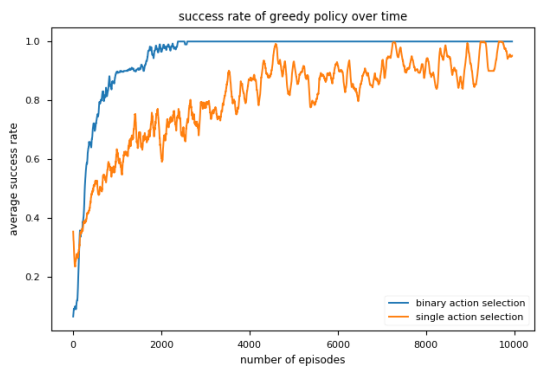}
    \caption{A comparison of learning rate for the standard DQN with a binary action selection, and an action selection that involves adding only a single node or edge per timestep.}
    \label{fig:actions}
\end{figure}

Figure \ref{fig:success} shows selected results of an implementation of this method. All data is averaged over 20 runs and training takes place over 10,000 episodes. For each run, a new ``ground truth" DAG is created for the given graph size and number of types, chosen at random. This is in order to prevent any bias from the selection of the graph. Figure~\ref{fig:actions} shows that a binary action selection results in a faster convergence rate, because it accounts for symmetries in the addition of edges. 

 Changes in total accumulated reward due to increasing the scale and node types of the DAGs is shown in Table \ref{sample-table} --- this decrease in accumulated reward is to be expected due to the exponentially-increasing size of the action space. For a DAG of only 6 nodes with only a single node type, there are 9,765 possible terminating states, of which approximately 5 are isomorphic to the `true' DAG that the agent is attempting to learn. Only these 5 states provide any non-zero reward, creating a highly sparse reward environment. For a DAG with 7 nodes, there are 615,195 possible graphs, again with the same amount of true graphs that provide any reward. It is clear then that learning over a span of 10,000 episodes is unlikely to learn anything, as it is highly unlikely any of these episodes will return any positive reward.

\begin{figure}[]
    \centering
    \includegraphics[width=.65\linewidth]{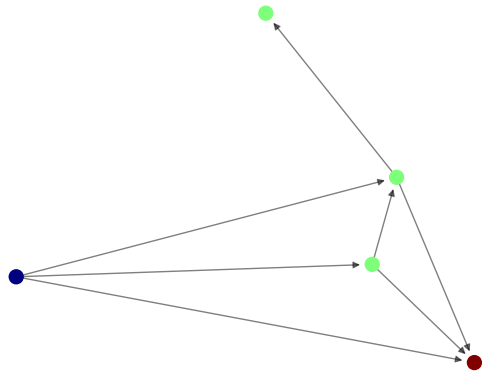}
    \caption{An example DAG with 5 nodes of 3 types. For the DAG to be considered isomorphic the nodes of each type need to be connected in the same manner. For this DAG there are 25,515 possible terminating states, of which only 3 return a non-zero reward.}
    \label{fig:my_label}
\end{figure}

\section{Conclusion}
\label{conclusion_section}
This paper introduced a deep Q-learning method for directed acyclic graph generation. By using a simple graph convolutional network based on spatial approaches, we can produce Q-values for various DAG states, with the transpose of the adjacency matrix used to account for edge directions. Only smaller graphs were tested in this implementation, due to the exponential size of the action space and the increased sparsity of the rewards.

For future work, a graph network that accounts for edge weights and continuous node features needs to be developed. A method that utilizes hierarchical reinforcement learning could significantly increase the convergence rate for DAGs at scale by learning structures of subgraphs found in generated graphs. Further improvements to the base DQN could also be implemented, in particular multi-step return calculations and dueling Q-networks as discussed in \cite{hessel2017rainbow}. Finally an application to a specific problem domain would allow for a more detailed reward design, and a more accurate test of this method.

\section*{Acknowledgements}

This research was sponsored by the U.S. Army Research Laboratory and the U.K. Ministry of Defence under Agreement Number W911NF-16-3-0001. The views and conclusions contained in this document are those of the authors and should not be interpreted as representing the official policies, either expressed or implied, of the U.S. Army Research Laboratory, the U.S. Government, the U.K. Ministry of Defence or the U.K. Government. The U.S. and U.K. Governments are authorized to reproduce and distribute reprints for Government purposes notwithstanding any copyright notation hereon.

\input{ms.bbl}
\end{document}

%% file: mathcommands.tex
\usepackage{amsmath,amsfonts,bm}

\def\eqref#1{equation~\ref{#1}}

\def\1{\bm{1}}

\def\rmB{{\mathbf{B}}}

\def\rmF{{\mathbf{F}}}

\def\rmH{{\mathbf{H}}}

\def\rmW{{\mathbf{W}}}

\def\vh{{\bm{h}}}

\def\mA{{\bm{A}}}

\def\mF{{\bm{F}}}

\def\mH{{\bm{H}}}

\DeclareMathAlphabet{\mathsfit}{\encodingdefault}{\sfdefault}{m}{sl}
\SetMathAlphabet{\mathsfit}{bold}{\encodingdefault}{\sfdefault}{bx}{n}

\def\gG{{\mathcal{G}}}

\newcommand{\normltwo}{L^2}


%% file: ms.bbl
\begin{thebibliography}{9}
\providecommand{\natexlab}[1]{#1}
\providecommand{\url}[1]{\texttt{#1}}
\expandafter\ifx\csname urlstyle\endcsname\relax
  \providecommand{\doi}[1]{doi: #1}\else
  \providecommand{\doi}{doi: \begingroup \urlstyle{rm}\Url}\fi

\bibitem[{De Cao} \& {Kipf}(2018){De Cao} and {Kipf}]{MolGAN}
{De Cao}, N. and {Kipf}, T.
\newblock {MolGAN: An implicit generative model for small molecular graphs}.
\newblock \emph{arXiv e-prints}, art. arXiv:1805.11973, May 2018.
\newblock URL \url{https://arxiv.org/abs/1805.11973}.

\bibitem[Hamilton et~al.(2017)Hamilton, Ying, and Leskovec]{GraphSAGE}
Hamilton, W.~L., Ying, R., and Leskovec, J.
\newblock Inductive representation learning on large graphs.
\newblock \emph{CoRR}, abs/1706.02216, 2017.
\newblock URL \url{https://arxiv.org/abs/1706.02216}.

\bibitem[Hessel et~al.(2017)Hessel, Modayil, van Hasselt, Schaul, Ostrovski,
  Dabney, Horgan, Piot, Azar, and Silver]{hessel2017rainbow}
Hessel, M., Modayil, J., van Hasselt, H., Schaul, T., Ostrovski, G., Dabney,
  W., Horgan, D., Piot, B., Azar, M., and Silver, D.
\newblock Rainbow: Combining improvements in deep reinforcement learning, 2017.
\newblock URL \url{https://arxiv.org/abs/1710.02298}.

\bibitem[Keskar et~al.(2016)Keskar, Mudigere, Nocedal, Smelyanskiy, and
  Tang]{batchorno}
Keskar, N.~S., Mudigere, D., Nocedal, J., Smelyanskiy, M., and Tang, P. T.~P.
\newblock On large-batch training for deep learning: Generalization gap and
  sharp minima.
\newblock \emph{CoRR}, abs/1609.04836, 2016.
\newblock URL \url{https://arxiv.org/abs/1609.04836}.

\bibitem[Orhean et~al.(2018)Orhean, Pop, and Raicu]{SchedulingQLearning}
Orhean, A.~I., Pop, F., and Raicu, I.
\newblock New scheduling approach using reinforcement learning for
  heterogeneous distributed systems.
\newblock \emph{J. Parallel Distrib. Comput.}, 2018.
\newblock URL \url{https://doi.org/10.1016/j.jpdc.2017.05.001}.

\bibitem[Schaul et~al.(2015)Schaul, Quan, Antonoglou, and
  Silver]{schaul2015prioritized}
Schaul, T., Quan, J., Antonoglou, I., and Silver, D.
\newblock Prioritized experience replay, 2015.
\newblock URL \url{https://arxiv.org/abs/1511.05952}.

\bibitem[Steenfatt et~al.(2018)Steenfatt, Nikolentzos, Vazirgiannis, and
  Zhao]{struc2vec++}
Steenfatt, N., Nikolentzos, G., Vazirgiannis, M., and Zhao, Q.
\newblock Learning structural node representations on directed graphs.
\newblock In \emph{Complex Networks and Their Applications {VII} - Volume 2
  Proceedings The 7th International Conference on Complex Networks and Their
  Applications {COMPLEX} {NETWORKS} 2018}, pp.\  132--144, 2018.
\newblock \doi{10.1007/978-3-030-05414-4\_11}.
\newblock URL \url{https://doi.org/10.1007/978-3-030-05414-4\_11}.

\bibitem[Xu et~al.(2018)Xu, Hu, Leskovec, and Jegelka]{howpowerfularegraphnn}
Xu, K., Hu, W., Leskovec, J., and Jegelka, S.
\newblock How powerful are graph neural networks?
\newblock \emph{CoRR}, abs/1810.00826, 2018.
\newblock URL \url{http://arxiv.org/abs/1810.00826}.

\bibitem[You et~al.(2018)You, Liu, Ying, Pande, and Leskovec]{RLDrugDiscovery}
You, J., Liu, B., Ying, R., Pande, V.~S., and Leskovec, J.
\newblock Graph convolutional policy network for goal-directed molecular graph
  generation.
\newblock \emph{CoRR}, abs/1806.02473, 2018.
\newblock URL \url{http://arxiv.org/abs/1806.02473}.

\end{thebibliography}
